\pdfoutput=1

\documentclass[conference]{IEEEtran}
\IEEEoverridecommandlockouts
\usepackage{cite}
\usepackage{amsmath,amssymb,amsfonts}
\usepackage{algorithmic}
\usepackage{graphicx}
\usepackage{textcomp}
\usepackage{xcolor}
\def\BibTeX{{\rm B\kern-.05em{\sc i\kern-.025em b}\kern-.08em
    T\kern-.1667em\lower.7ex\hbox{E}\kern-.125emX}}
    
\newtheorem{defn}{Definition}
\usepackage{algorithm}
\usepackage{caption}
\usepackage{subfig}
    
\usepackage[pscoord]{eso-pic}
\usepackage{hyperref}
\newcommand{\placetextbox}[3]{
  \setbox0=\hbox{#3}
  \AddToShipoutPictureFG*{
    \put(\LenToUnit{#1\paperwidth},\LenToUnit{#2\paperheight}){\vtop{{\null}\makebox[0pt][c]{#3}}}%
  }%
}%

\begin{document}

\placetextbox{0.5}{0.998}{\normalfont \small \textcopyright 2020 IEEE. Personal use of this material is permitted. Permission from IEEE must be obtained for all other uses, in any current}

\placetextbox{0.5}{0.988}{\normalfont \small or future media, including reprinting/republishing this material for advertising or promotional purposes, creating new collective}

\placetextbox{0.5}{0.978}{\normalfont \small works, for resale or redistribution to servers or lists, or reuse of any copyrighted component of this work in other works}

\placetextbox{0.5}{0.967}{\normalfont \small Author accepted manuscript, published in ``IEEE 2022 International Joint Conference on Neural Networks (IJCNN), 2022, pp. 1-8''.}

\placetextbox{0.5}{0.952}{\normalfont DOI: \url{https://doi.org/10.1109/IJCNN55064.2022.9892809}.}%

\title{Analysis of Trade-offs in Fair Principal Component Analysis Based on Multi-objective Optimization\\

\thanks{The authors thank the grants \#2021/11086-0, \#2020/10572-5, \#2020/09838-0, \#2020/01089-9 and \#2019/20899-4, São Paulo Research Foundation (FAPESP), and the grants \#311357/2017-2, National Council for Scientific and Technological Development (CNPq), for the financial support.}
}


\author{\IEEEauthorblockN{Guilherme D. Pelegrina}
\IEEEauthorblockA{\textit{School of Applied Sciences} \\
\textit{University of Campinas}\\
Limeira, Brazil \\
guidean@unicamp.br}
\and
\IEEEauthorblockN{Renan D. B. Brotto}
\IEEEauthorblockA{\textit{School of Electrical and Computer Engineering} \\
\textit{University of Campinas}\\
Campinas, Brazil \\
rbrotto@decom.fee.unicamp.br}
\and
\IEEEauthorblockN{Leonardo T. Duarte}
\IEEEauthorblockA{\textit{School of Applied Sciences} \\
\textit{University of Campinas}\\
Limeira, Brazil \\
leonardo.duarte@fca.unicamp.br}
\and
\IEEEauthorblockN{Romis Attux}
\IEEEauthorblockA{\textit{School of Electrical and Computer Engineering} \\
\textit{University of Campinas}\\
Campinas, Brazil \\
attux@dca.fee.unicamp.br}
\and
\IEEEauthorblockN{João M. T. Romano}
\IEEEauthorblockA{\textit{School of Electrical and Computer Engineering} \\
\textit{University of Campinas}\\
Campinas, Brazil \\
romano@dmo.fee.unicamp.br}
}

\maketitle

\begin{abstract}
In dimensionality reduction problems, the adopted technique may produce disparities between the representation errors of different groups. For instance, in the projected space, a specific class can be better represented in comparison with another one. In some situations, this unfair result may introduce ethical concerns. Aiming at overcoming this inconvenience, a fairness measure can be considered when performing dimensionality reduction through Principal Component Analysis. However, a solution that increases fairness tends to increase the overall reconstruction error. In this context, this paper proposes to address this trade-off by means of a multi-objective-based approach. For this purpose, we adopt a fairness measure associated with the disparity between the representation errors of different groups. Moreover, we investigate if the solution of a classical Principal Component Analysis can be used to find a fair projection. Numerical experiments attest that a fairer result can be achieved with a very small loss in the overall reconstruction error.
\end{abstract}


\section{Introduction}
Machine Learning (ML) techniques have been highlighted in the last years for its broad scope of application in many different areas, such as pattern recognition~\cite{Duda2000,Bishop2006}, signal processing~\cite{Little2019}, audio and image processing~\cite{Camastra2015}. Besides its ubiquitous use in technical problems, ML models have also been applied in tasks with social and economical impacts, e.g., credit concession~\cite{Hardt2016}, recidivism prediction~\cite{Chouldechova2017} and gerrymandering~\cite{Kearns2018}. Although these models can achieve a good performance, generally, they are not conceived in order to incorporate ethical concerns. As a consequence, unfair ML techniques may lead to discrimination problems~\cite{Barocas2019,Kearns2019}. Therefore, apart from the performance measure, the ML models must also incorporate additional information in the ML model construction, such as a fairness measure~\cite{Samadi2018,Hwang2020}.

The fairness concern can be addressed in the preprocessing \cite{Wang2018,Samadi2018}, model training~\cite{Zafar2019} or post-processing~\cite{Lohia2019} steps. For instance, in the preprocessing step, which is the focus of this paper, fairness is generally associated with how one manages the acquired dataset. As presented in~\cite{Samadi2018}, the dimensionality reduction provided by the classical Principal Components Analysis (PCA)~\cite{Jolliffe2002} may be biased towards one group of the population, leading to different reconstruction errors. As a consequence, the application of PCA may lead to disparate performances when this data is used in a ML model.

Motivated by the aforementioned ethical concerns, the first goal of this work is to introduce a fairness measure when dealing with dimensionality reduction using PCA. More specifically, alongside the overall reconstruction error (the reconstruction error when using all samples, without a division into groups), we also consider a fairness measure as a cost function when seeking for the projection matrix. However, differently from existing works, which incorporate both objectives into a single one~\cite{Samadi2018}, we address this problem in a multi-objective fashion~\cite{Miettinen1999}. Therefore, in our formulation, called Multi-Objective Fair Principal Component Analysis (MOFPCA), both criteria are optimized simultaneously. Moreover, instead of searching for a new projection matrix, we use the solution of the classical PCA to extract the projection vectors that will be used to optimize both objectives. As a result, we obtain a set of solutions that can be used to visualize the trade-offs between the considered objectives. This may be helpful in practical applications, since it provides a transparent way to support the decision under these conflicting objectives. Moreover, with this framework, one may also verify if there is a single solution associated with a small value of the reconstruction error and a fair representation of both classes.

The paper is organized as follows. In Section~\ref{sec:related_work}, we describe the existing works related to our proposal. We discuss the addressed problem in Section~\ref{sec:probform}. Sections~\ref{sec:moo} and~\ref{sec:prop} present the multi-objective optimization formulation and the proposed MOFPCA approach, respectively. The numerical experiments are presented in Section~\ref{sec:exper} and, finally, the concluding remarks and future perspectives are described in Section~\ref{sec:concl}.

\section{Related works}
\label{sec:related_work}

In \cite{Olfat2019}, the authors presented a fair dimensionality reduction method based on a Semi Definite Program for PCA and Kernel PCA. They tackled a classification problem where the disparate impact was used as a fairness measure. Fairness in PCA was also conducted in~\cite{Samadi2018}, where the authors proposed a fairness criteria based on the loss suffered from each group in the projected space with respect to their individual optimal projection. They start by independently performing PCA for each protected group. In each execution of PCA (on all samples), the reconstruction error for each group is calculated and the loss function indicates how far from the two benchmarks the obtained values lies. In the optimal scenario, the proposed \textit{FairPCA} algorithm leads to a projection that deviates equally each population from their ideal reconstruction error. Recently,~\cite{Zalcberg2021} addressed fairness in PCA by means of a sub-gradient descent algorithm and provided an analysis of this problem when the target vectors are orthogonal to each other.

In the context of multi-objective optimization, fairness in PCA has also been addressed in~\cite{Kamani2019}. The authors proposed a gradient descent algorithm for the Fair PCA projection matrix. Moreover, the concept of non-dominance is based on the gradient direction and, therefore, only considers first order information. This may lead to dominated solutions in more general scenarios (where the cost function may not be convex) and it is necessary to use more robust algorithms to explore the feature space.

A common point on the aforementioned works is that they search for a projection matrix that is different from the one obtained by the classical PCA. On the other hand, our proposal aims at using the principal components already obtained from the classical PCA and then sorting them to attain the fairness concern. In other words, with the purpose of reducing the number of attributes from $d$ to a $r$-dimensional space, we combine $r$ directions of the classical PCA differently from the solution that minimizes only the total reconstruction error. Therefore, the multi-objective proposed framework acts in the selection of the components that optimizes both reconstruction error and fairness measure simultaneously. This characteristic is interesting in already running data systems, since it introduces a re-sorting block that can be easily plugged into the data pipeline. Moreover, it takes advantage from the efficient algorithms to perform PCA even with high dimensional data~\cite{Fan2018}.

\section{Disparities in PCA}
\label{sec:probform}

Consider a data matrix $\mathbf{X} \in \mathbb{R}^{n \times d}$, with $n$ as the number of $d$-dimensional samples. In summary, the aim of Principal Components Analysis~\cite{Jolliffe2002} is to seek for an orthogonal projection matrix $\mathbf{U} \in \mathbb{R}^{d \times r}$ that reduces the data dimension while minimizing the (overall) reconstruction error $\mathcal{R}(\mathbf{U})$, given by
\begin{equation}
\mathcal{R}(\mathbf{U}) = \frac{1}{n}\left\|\mathbf{X} - \mathbf{XUU}^T\right\|_{F}^{2},
\label{eq:recons_error}
\end{equation}
where $\left\| \cdot \right\|_{F}$ is the Frobenius norm~\cite{Golub2013}. Mathematically, the optimization problem tackled by PCA can be expressed by:
\begin{equation}
\begin{array}{ll}
    \underset{\mathbf{U}}{\min} & \mathcal{R}(\mathbf{U}) \\
		\text{s.t.} & \mathbf{U}^T\mathbf{U} = \mathbf{I},
\end{array}
\label{eq:classical_PCA}
\end{equation}
where $\mathbf{I}$ is the identity matrix. Without loss of generality, let us consider that the first $l$ columns of $\mathbf{U}$ represent the first $l$ principal components. Clearly, the PCA formulation expressed in~\eqref{eq:classical_PCA} does not consider possible disparities between different groups. Aiming at overcoming unfair results in dimensionality reduction, it is fundamental to consider an adequate fairness measure. In this paper, we directly associate this measure with the reconstruction errors for individual classes. In order to formulate this criterion, let us assume that the samples can be divided as $\mathbf{X} = \left[ \mathbf{X}_A; \mathbf{X}_B \right]$, where $\mathbf{X}_A \in \mathbb{R}^{n_A \times d}$ and $\mathbf{X}_B \in \mathbb{R}^{n_B \times d}$ represent different groups (e.g., male and female). The proposed fairness measure is defined by
\begin{equation}
    \mathcal{F}\left(\mathbf{U} \right) = \left( \frac{\left\|\mathbf{X}_A - \mathbf{X}_A\mathbf{U}\mathbf{U}^T \right\|_{F}^{2}}{n_A}  - \frac{\left\|\mathbf{X}_B - \mathbf{X}_B\mathbf{U}\mathbf{U}^T \right\|_{F}^{2}}{n_B} \right)^2.
    \label{eq:fairness_measure}
\end{equation}
In short, the idea in~\eqref{eq:fairness_measure} is to evaluate the disparity between the reconstruction errors for classes $\mathbf{X}_A$ and $\mathbf{X}_B$. Ideally, in the fairest projection, matrix $\mathbf{U}$ is the one leading to $\mathcal{F}(\mathbf{U}) = 0$.

In this paper, we consider that both total reconstruction error and fairness measure are equally important in the dimensionality reduction problem. Therefore, both cost functions should be optimized simultaneously. For this purpose, we consider the application of multi-objective optimization to deal with fairness in PCA. We address this technique in the next section.

\section{Multi-objective optimization}
\label{sec:moo}

In mono-objective optimization, the solution is the feasible one that optimizes a single cost function. However, in multi-objective optimization, the notion of optimality must be extended to a vector-valued cost function. In our case, the optimal solution should be the one that leads to minimum values of $\mathcal{J} = \left[\mathcal{R}(\mathbf{U}^{*}),\mathcal{F}(\mathbf{U}^{*}) \right]$. Very often multi-objective optimization problems involve conflicting cost functions. Therefore, one rarely finds a single solution that optimizes all objectives simultaneously. In the addressed problem, it is expected a conflict between the adopted cost functions, i.e., a solution that minimizes the overall reconstruction error may not be the fairest one (and vice-versa). Although this compromise exists, we may have a set of solutions that achieve suitable performances on both objectives. The definition of such solutions is based on the following concept of dominance~\cite{Deb2001,Miettinen1999}:

\begin{defn}{\textbf{(Dominance):}}
\label{def:domin}
Consider two feasible solutions $\mathbf{\tilde{U}}$ and $\mathbf{\tilde{U}}'$. We say that $\mathbf{\tilde{U}}$ dominates $\mathbf{\tilde{U}}'$ if $\mathbf{\tilde{U}}$ is as good as $\mathbf{\tilde{U}}'$ in all objectives and $\mathbf{\tilde{U}}$ is strictly better than $\mathbf{\tilde{U}}'$ in at least one objective.
\end{defn}

Based on this concept, we say that a solution is non-dominated if none of the objectives (reconstruction error and fairness measure, in our case) can be improved without degrading the other one (i.e., if a solution leads to a better reconstruction error, it must lead, at the same time, to a worse fairness measure and vice-versa). If we assume that both Equations~\eqref{eq:recons_error} and~\eqref{eq:fairness_measure} should be minimized, the solutions of the multi-objective optimization problem (also called non-dominated set or Pareto front~\cite{Deb2001,Miettinen1999}) are the following:

\begin{defn}{\textbf{(Non-dominated solution):}}
\label{def:nondomin}
A solution $\mathbf{\tilde{U}}$ is a non-dominated solution if there is no other solution $\mathbf{\tilde{U}}'$ such that (i) both $\mathcal{R}(\mathbf{\tilde{U}}') \leq \mathcal{R}(\mathbf{\tilde{U}})$ and $\mathcal{F}(\mathbf{\tilde{U}}') \leq \mathcal{F}(\mathbf{\tilde{U}})$ and (ii) either $\mathcal{R}(\mathbf{\tilde{U}}') < \mathcal{R}(\mathbf{\tilde{U}})$ or $\mathcal{F}(\mathbf{\tilde{U}}') < \mathcal{F}(\mathbf{\tilde{U}})$.
\end{defn}

\section{Proposed approach}
\label{sec:prop}

In this paper, we propose a multi-objective-based approach for Fair PCA. Moreover, for a dimensionality reduction problem from $d$ to $r$-dimensional space, we use the solution of the classical PCA and select $r$ principal components that optimize both Equations~\eqref{eq:recons_error} and~\eqref{eq:fairness_measure} simultaneously. Assume $\mathbf{q}^{*} = \left[ q_1^{*}, \ldots, q_r^{*} \right]$, where each $q_i^{*} \in \left\{1, 2, \ldots, d \right\}$, $i=1, \ldots, r$ indicates the selected principal components (columns of $\mathbf{U}$). The multi-objective optimization problem is expressed by
\begin{equation}
\begin{aligned}
    \underset{[q_1^{*}, \ldots, q_r^{*}]}{\min} \, \, \, \left[\mathcal{R}(\mathbf{U}^{*}),\mathcal{F}(\mathbf{U}^{*}) \right],
\end{aligned}
\label{eq:fair_PCA}
\end{equation}
where $\mathbf{U}^{*} = \left[ \mathbf{u}_{q_1^{*}}, \mathbf{u}_{q_2^{*}}, \ldots, \mathbf{u}_{q_r^{*}} \right]$ is the adjusted projection matrix whose columns are composed by the principal components extracted from $\mathbf{U}$. Note that $\mathbf{u}_{q_1^{*}}, \mathbf{u}_{q_2^{*}}, \ldots, \mathbf{u}_{q_r^{*}}$ are not, necessarily, the first $r$ columns of $\mathbf{U}$. For example, if $\mathbf{q}^{*} = \left[ 1, 3, 6 \right]$ (i.e, the columns 1, 3 and 6 of $\mathbf{U}$ are a solution of~\eqref{eq:fair_PCA}), then $\left[ \mathbf{u}_{q_1^{*}}, \mathbf{u}_{q_2^{*}}, \mathbf{u}_{q_3^{*}} \right] = \left[\mathbf{u}_1, \mathbf{u}_3, \mathbf{u}_6 \right]$. Moreover, one does not need to include an orthogonality constraint on $\mathbf{U}^{*}$, since the selected columns already ensure this property. Therefore, one only needs to select a combination of the principal components provided by the classical PCA.

There are several existing methods that deal with multi-objective optimization problems~\cite{Miettinen1999}. In this paper, we tackle the MOFPCA formulation expressed in~\eqref{eq:fair_PCA} by means of evolutionary computation~\cite{Deb2001,Zhou2011}, which has been used in several ML applications~\cite{Chen2010,Mukhopadhyay2014a,Mukhopadhyay2014b,Canuto2018,Fernandes2020,He2020}. With respect to the adopted algorithm, we considered an improved version of the Strength Pareto Evolutionary Algorithm (SPEA)~\cite{Zitzler1998}, called SPEA2~\cite{Zitzler2001}. We adopted the SPEA2 since we deal with a multi-objective combinatorial problem and, as mentioned in~\cite{Ehrgott2003}, this algorithm can be used to deal with such situations. Moreover, it has been applied in ML tasks~\cite{Dalip2014,DeSousa2016,Canuto2018}.

Although the implementation of SPEA2 is quite simple, one generally needs to set some specific concepts, as described in the sequel. We also highlight the particularities of each one (for more details, see~\cite{Zitzler2001}).

\begin{itemize}
\item \textbf{Individual}: An individual $\mathbf{q}$ leads to a possible solution for the multi-objective problem. In our proposal, since we search for a projection matrix $\mathbf{U}^{*}$ whose columns are composed by $r$ principal components extracted from the classical PCA (from a total of $d$), an individual comprises a combination of $r$ coefficient indices. For example, for a projection from $d=10$ to $r=3$-dimensional space, an individual could be $\mathbf{q} = \left[1,3,6 \right]$, which leads to a projection matrix $\mathbf{\tilde{U}} = \left[ \mathbf{\tilde{u}}_{q_1}, \mathbf{\tilde{u}}_{q_2}, \mathbf{\tilde{u}}_{q_3}, \mathbf{\tilde{u}}_{q_4} \right] = \left[ \mathbf{u}_1, \mathbf{u}_3, \mathbf{u}_3 \right]$. 
\item \textbf{Population}: The population $\mathbf{P}$ is composed by a set of individuals. The size of population, represented by $\bar{P}$ is predefined before the algorithm starts. Moreover, in order to define the initial population, we randomly generated $\bar{P} - 1$ individuals. In this procedure, aiming at covering a large region of the feasible space, we also ensured that they are different. For the other individual, we set $\mathbf{q} = \left[1, \ldots, r \right]$, which is the optimal solution for the overall reconstruction error. Therefore, this may speed-up the convergence.
\item \textbf{External set}: Another element used in SPEA2 is the external set, represented by $\mathbf{E}$. In the beginning of the algorithm, the external set is empty. However, after each generation (or iteration) $g$, it is updated with the non-dominated solutions found so far. Similarly as in the population, we also predefine the external set size $\bar{E}$.
\item \textbf{Fitness}: Based on both $\mathcal{R}(\mathbf{\tilde{U}})$ and $\mathcal{F}(\mathbf{\tilde{U}})$, the fitness value indicates how good (or bad) is the performance of an individual $\mathbf{q}$ in terms of dominance. Therefore, based on this measure, we can rank the individuals and select the best ones after each generation. For more details the fitness measure calculation in SPEA2, see~\cite{Zitzler2001}.
\item \textbf{Crossover and mutation}: Both crossover and mutation are evolutionary operators used to generate new individuals. The crossover consists in generating a new individual $\mathbf{q}$, which will be used in generation $g+1$, by randomly taking parts of two different individuals of generation $g$. In mutation, we create a new individual used in generation $g+1$ by randomly modifying a part (some indices, in our case) of an individual of generation $g$. An import aspect in the addressed problem associated with both operators is that we must ensure the feasibility of the new created individuals. In our proposal, when applying both operators, we must verify that there is no repetition in the indices that compose each individual.
\end{itemize}

With the aforementioned concepts been clarified, the steps of SPEA2 are presented in Algorithm~\ref{alg:spea}. As inputs, we predefine the initial population $\mathbf{P}$, the (initially) empty external set $\mathbf{E}$, the population size $\bar{P}$, the external set size $\bar{E}$, the maximum number of iterations $G$ and the crossover rate $\alpha$. In the first step, we calculate the fitness values for all individuals in both population and external set. These values are used in Step 2 in order to select the best individuals found so far and update the external set. At this point, if we achieve the maximum number of iterations, we set $\mathbf{E}$ as the non-dominated solutions and stop the algorithm. Otherwise, we move to the mating selection step (Step 3). In this step, we perform a binary tournament selection on the external set in order to define the set of individuals that will be submitted to the evolutionary operators in the variation step (Step 5). It is worth mentioning that, in Step 5, $\alpha$\% of the new population $\mathbf{P}$ is obtained through crossover and the other $\left(1 - \alpha \right)$\% through mutation. The variation step ends by setting $g = g + 1$ and the algorithm return to Step 1.

\begin{algorithm}
    \caption{SPEA2}
    \label{alg:spea}
		\begin{algorithmic}
				\STATE \textbf{Input:} $\mathbf{P}$, $\mathbf{E}$, $\bar{P}$, $\bar{E}$, $G$ and $\alpha$.
				\STATE \textbf{Output:} External set $\mathbf{E}$.
				\STATE Set $g=1$.
				\WHILE{$g \leq G$}
						\STATE \textit{Step 1: Fitness assignment}. For each individual in $\mathbf{P}$ and $\mathbf{E}$, calculate the fitness measure.
						\STATE \textit{Step 2: Selection}. Based on the fitness values, select the best $\bar{E}$ individuals and update $\mathbf{E}$.
						\IF{$g = G$}
								\STATE \textit{Step 3: Termination}. Define the external set $\mathbf{E}$ as the non-dominated set and stop the algorithm.
						\ENDIF
						\STATE \textit{Step 4: Mating selection}. Select $\bar{P}$ individuals for the next step through a binary tournament selection in $\mathbf{E}$.
						\STATE \textit{Step 5: Variation}. Apply crossover and mutation on the individuals selected on the previous step and generate the new population $\mathbf{P}$. Set $g = g + 1$.
				\ENDWHILE
    \end{algorithmic}
\end{algorithm}

When SPEA2 finishes, we achieve a set of non-dominated solutions. In terms of Pareto optimality, they are equally optimal. Therefore, one cannot say that a specific non-dominated solution is better than other one by taking into account the considered reconstruction error and fairness measure. In other words, we are not able to select a single optimal solution that solves our MOFPCA problem. However, an interesting aspect of the multi-objective approach is that we can visualize the compromise between the non-dominated solutions. This enhances the transparency the the ML model, since the users can see how much they are willing to lose in the reconstruction error in order to improve fairness. 

\subsection{On the selection of a non-dominated solution}
\label{subsec:selec}

\begin{figure}[ht]
\centering
\includegraphics[height=2.2cm]{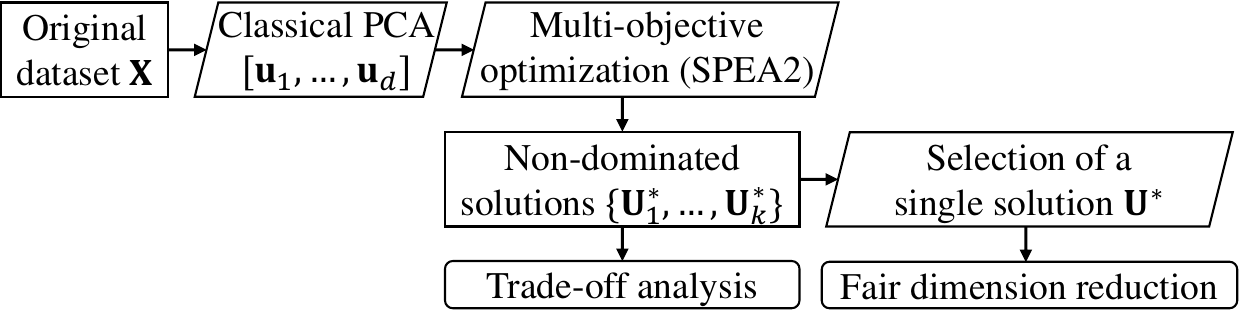}
\caption{The proposed MOFPCA scheme.}
\label{fig:framework}
\end{figure}

Although all solutions in the non-dominated set are equally optimal, in this paper, we also propose a technique to select a single one. Assume that $\mathbf{\Omega}$ represents the set of non-dominated solutions. Moreover, assume that $\mathbf{r}$ and $\mathbf{f}$ represent the set of reconstruction errors and fairness measures, respectively, obtained by all $\mathbf{q}^{*} \in \mathbf{\Omega}$. The proposed technique consists in selecting the $\mathbf{q}^{*} \in \mathbf{\Omega}$ (and, therefore, the $\mathbf{U}^{*}$) that minimizes the sum of the normalized reconstruction error and fairness measure. Mathematically, $\mathbf{U}^{*}$ is obtained by
\begin{equation}
\begin{array}{ll}
    \underset{\mathbf{q}^{*}}{\min} & \lambda \mathcal{\tilde{R}}(\mathbf{U}^{*}) + \left(1-\lambda \right)\mathcal{\tilde{F}}(\mathbf{U}^{*}) \\
		\text{s.t.} & \mathbf{q}^{*} \in \mathbf{\Omega},
\end{array}
\label{eq:select}
\end{equation}
where $\mathcal{\tilde{R}}(\mathbf{U}^{*}) = \frac{\mathcal{R}(\mathbf{U}^{*}) - \min(\mathbf{r} )}{\max(\mathbf{r} ) - \min(\mathbf{r} ) }$, $\mathcal{\tilde{F}}(\mathbf{U}^{*}) = \frac{\mathcal{F}(\mathbf{U}^{*}) - \min(\mathbf{f} ) }{\max(\mathbf{f} ) - \min(\mathbf{f} )}$ ($\min(\cdot )$ and $\max(\cdot )$ represent the minimum and maximum operators, respectively) and $\lambda$ is a predefined weighting factor that controls the importance given to each objective. For example, if one considers that fairness is more important than the overall reconstruction error, we would assume $\lambda \leq 0.5$. However, in our experiments, we assumed $\lambda = 0.5$, which means that both objectives are equally important. Therefore, the normalization will only compensate the difference between the scales. Figure~\ref{fig:framework} illustrates the MOFPCA scheme.

It is important to recall that the selection of a single solution within the non-dominated set is not mandatory to analyze the trade-off between the objectives. We here use this solution in order to compare the MOFPCA approach with both classical PCA and FairPCA algorithm~\cite{Samadi2018}.

\section{Numerical experiments}
\label{sec:exper}

In order to verify the application of the proposed MOFPCA in real dimensionality reduction scenarios, we consider two datasets: Default Credit~\cite{Yeh2009} and Labeled Faces in the Wild (LFW)~\cite{Huang2008}. We considered these datasets, as well as the adopted sensitive attributes, for the purpose of comparison, since some related works~\cite{Samadi2018,Kamani2019}, mentioned in Section~\ref{sec:related_work}, also used them in their experiments.


\subsection{Experiments with the Default Credit dataset}
\label{subsec:exper1}

The Default Credit dataset comprises $n=30,000$ samples and $m=23$ attributes. Among them, we adopt the education level as the sensitive one (and removed it from the dataset). Therefore, we have $n_1=5,385$ and $n_2=24,615$ samples associated with lower (high school and others) and higher (graduate school and university) education levels, respectively. All the attributes were normalized in order to have zero mean and unitary variance.

\begin{figure}[ht]
\centering
\includegraphics[height=5.7cm]{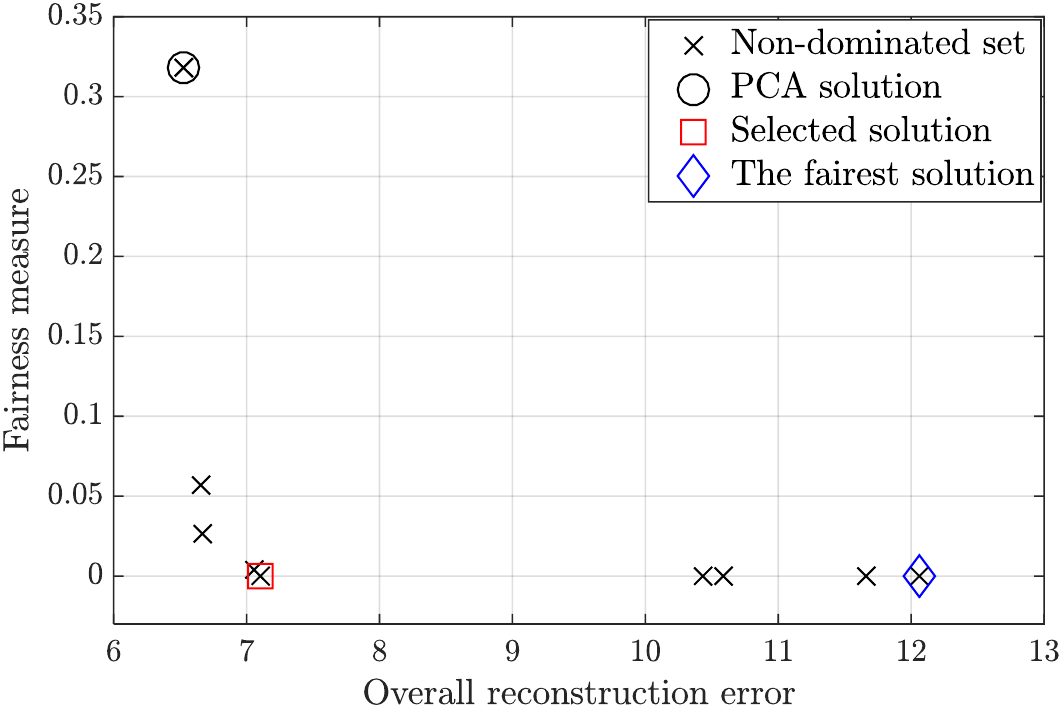}
\caption{Non-dominated solutions for 6 features - Default Credit dataset.}
\label{fig:pareto}
\end{figure}

In order to illustrate the non-dominated solutions achieved by the proposed MOFPCA approach\footnote{In all experiments conducted in this paper, the SPEA2 parameters were experimentally defined by $\alpha = 50$, $\bar{P} = \min\left(100,round\left(\frac{1}{2}\frac{d!}{r!(d-r)!}\right)\right)$ and $\bar{E} = round\left(\bar{P}/2\right)$, where $round(\cdot)$ returns the closest integer. With respect to the number of iterations, we adopted $G = 30$ and $G = 50$ for the Default Credit and the LFW datasets, respectively. We consider that this parameters setting led to a good algorithm convergence.}, let us reduce the number of features from $22$ to $6$-dimensional samples. The non-dominated set is presented in Figure~\ref{fig:pareto}. Note that, among the compromising solutions, we have the solution of a classical PCA (the one that minimizes the reconstruction error), the fairest projection (the one that minimizes the adopted fairness measure) and the selected one (as described in Section~\ref{subsec:selec}). In this case, the selected solution is composed by the principal components $\mathbf{u}_1, \mathbf{u}_2, \mathbf{u}_3, \mathbf{u}_5, \mathbf{u}_{7}$ and $\mathbf{u}_{8}$ of $\mathbf{U}$. Therefore, we can note that a simple change in the order of the principal components (e.g, by removing both $4$-th and $6$-th columns of $\mathbf{U}$ and by adding the $7$-th and $8$-th ones) significantly increased fairness (or decreased the fairness measure described in Equation~\eqref{eq:fairness_measure}) with a small loss in the overall reconstruction error.

\begin{figure}[!h]
\centering
\subfloat[Reconstruction error.]{\includegraphics[width=3.0in]{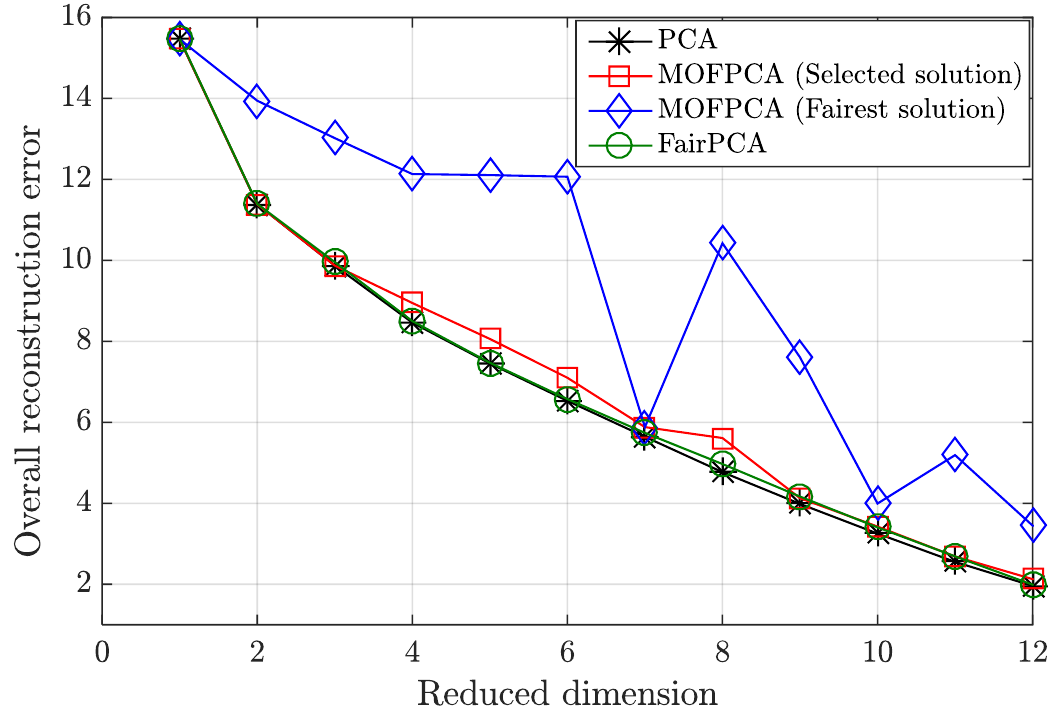}
\label{fig:result_recerror}}
\hfil
\subfloat[Fairness measure.]{\includegraphics[width=3.0in]{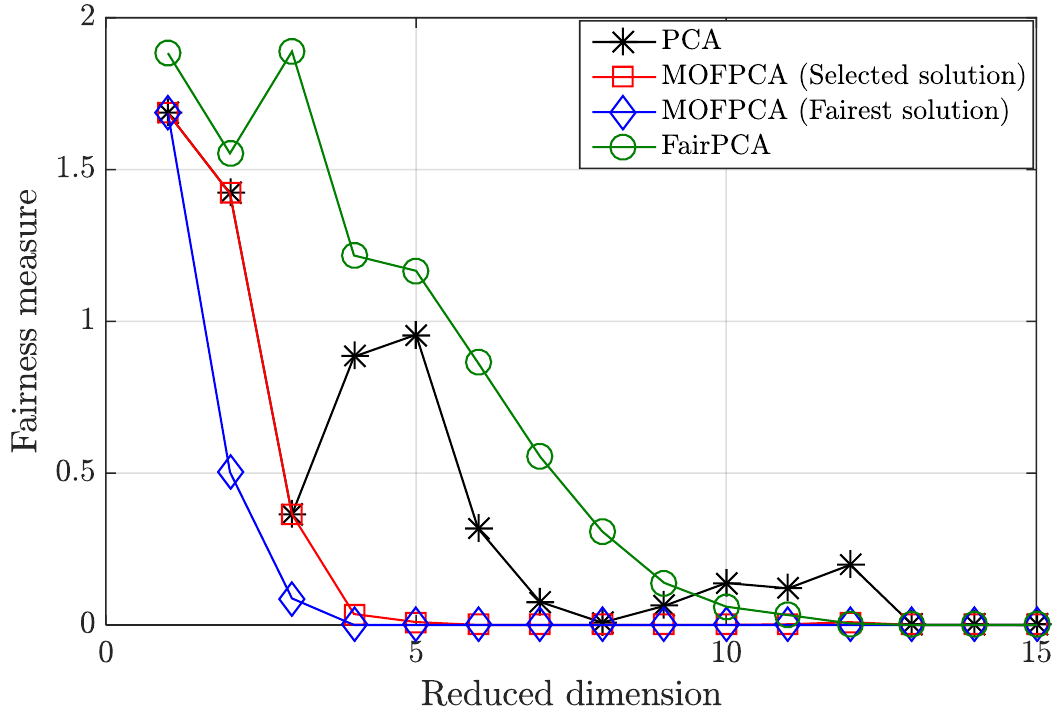}
\label{fig:result_fairmeas}}
\caption{Results for both cost functions and different numbers of reduced dimensions - Default Credit dataset.}
\label{fig:result_all}
\end{figure}

We also applied our proposal to different numbers of reduced dimensions. Figure~\ref{fig:result_all} presents the obtained overall reconstruction errors and fairness measures. We compared the results provided by PCA, the selected MOFPCA solution, the fairest solution and FairPCA~\cite{Samadi2018}. One may note in Figure~\ref{fig:result_recerror} that the solution selected by our proposal led to reconstruction errors very close to the ones obtained by both FairPCA and PCA, the latter being the benchmark for this cost function. However, in terms of the fairness measure, Figure~\ref{fig:result_fairmeas} indicates a relevant difference between the considered approaches: the MOFPCA led to a better fairness condition with fewer features than the PCA and FairPCA. Moreover, the performance of the selected solution was very close to the fairest one for at least 4 dimensions. Although FairPCA considers fairness in its formulation, it led to the worst adopted fairness measure for $r < 10$. It is worth recalling that the fairness measure considered in FairPCA (see~\cite{Samadi2018} for further details) is different from~\eqref{eq:fairness_measure}.

\begin{figure*}[!h]
\centering
\subfloat[PCA.]{\includegraphics[width=1.7in]{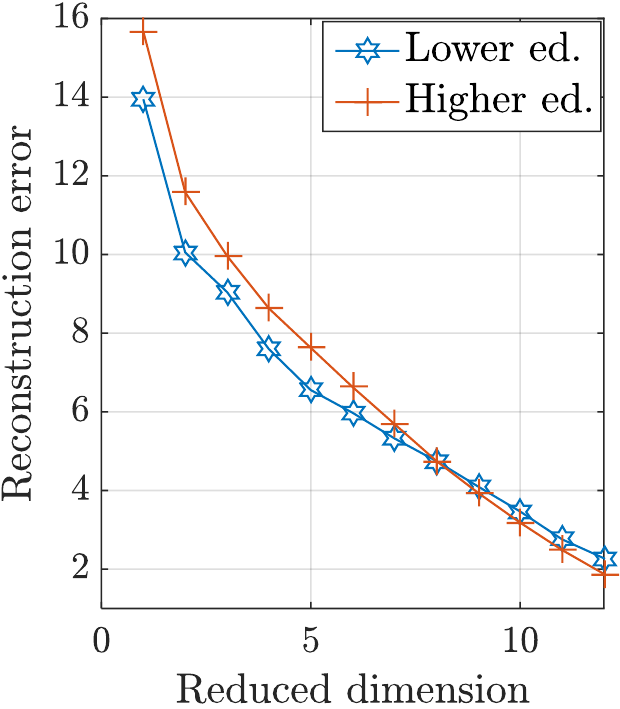}
\label{fig:reconserror_pca}}
\hfil
\subfloat[MOFPCA (Selected solution).]{\includegraphics[width=1.7in]{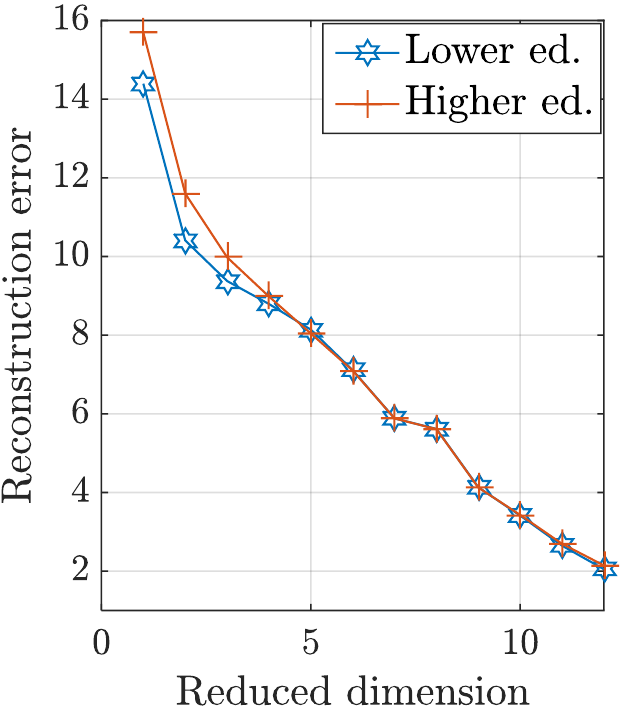}
\label{fig:reconserror_mofpca}}
\hfil
\subfloat[MOFPCA (Fairest solution).]{\includegraphics[width=1.7in]{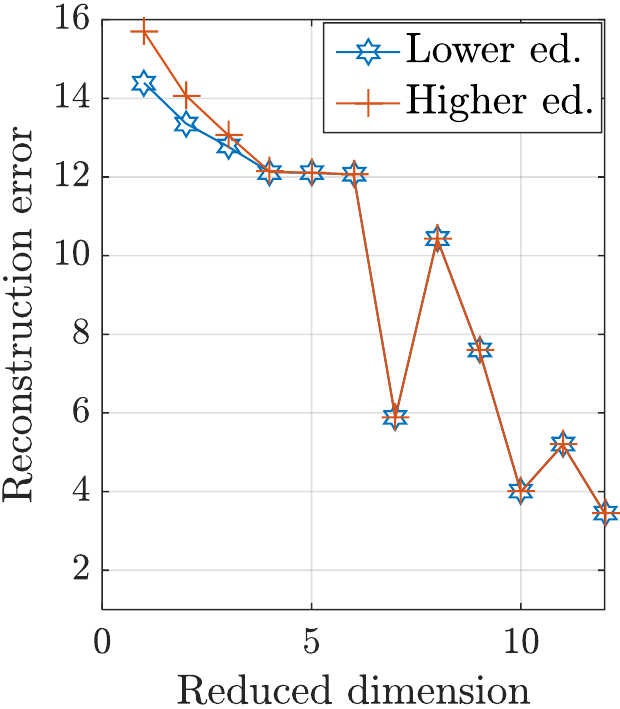}
\label{fig:reconserror_mofpca_f}}
\hfil
\subfloat[FairPCA.]{\includegraphics[width=1.7in]{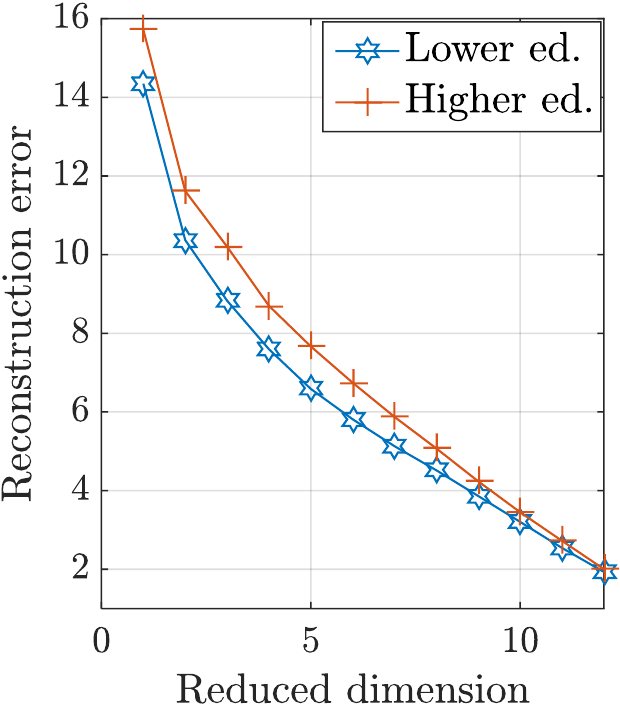}
\label{fig:reconserror_samadi}}
\caption{Reconstruction errors for lower and higher education levels - Default Credit dataset.}
\label{fig:reconserror_all}
\end{figure*}

The difference between the reconstruction errors for each class can be visualized in Figure~\ref{fig:reconserror_all}. The results in Figure~\ref{fig:reconserror_mofpca} attested that this difference, when applying our proposal, tends to zero for all dimensionality reduction with $r \ge 5$. On the other hand, in FairPCA (Figure~\ref{fig:reconserror_samadi}), this disparity is mitigated only for $r \geq 13$. If we consider the fairest solution, the reconstruction errors are very close for $r \geq 3$. However, we considerably increase the overall reconstruction error.

\subsection{Experiments with the LFW dataset}
\label{subsec:exper2}

The LFW dataset consists of $m=1764$ attributes (pixels) and $n=13,232$ samples, divided into two groups: female ($n_1=2,962$) and male ($n_2=10,270$). As mentioned in~\cite{Samadi2018}, the gender information was manually verified by~\cite{Afifi2017}. All pixels were normalized by $1/255$.

\begin{figure}[!ht]
\centering
\includegraphics[height=5.7cm]{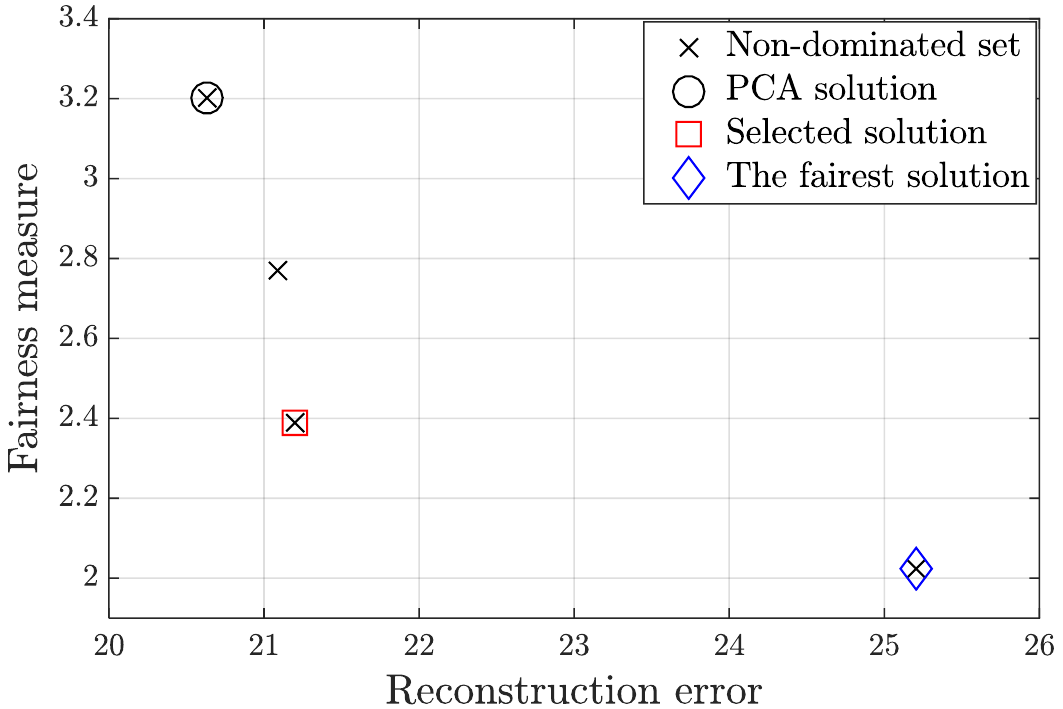}
\caption{Non-dominated solutions for 5 features - LFW dataset.}
\label{fig:pareto_images}
\end{figure}

By taking the projection into $r=5$ dimensions, the achieved non-dominated solutions are presented in Figure~\ref{fig:pareto_images}. Similarly as in the Default Credit dataset, we may also visualize here the trade-offs between the reconstruction error and the adopted fairness measure. In contrast with Figure~\ref{fig:pareto}, the selected solution, which is composed by the principal components $\mathbf{u}_1, \mathbf{u}_2, \mathbf{u}_3, \mathbf{u}_4$ and $\mathbf{u}_{8}$ of $\mathbf{U}$, does not lead to very good results in both cost functions. However, it could improve fairness with some loss in the overall reconstruction error.

\begin{figure}[ht]
\centering
\subfloat[Reconstruction error.]{\includegraphics[width=3.0in]{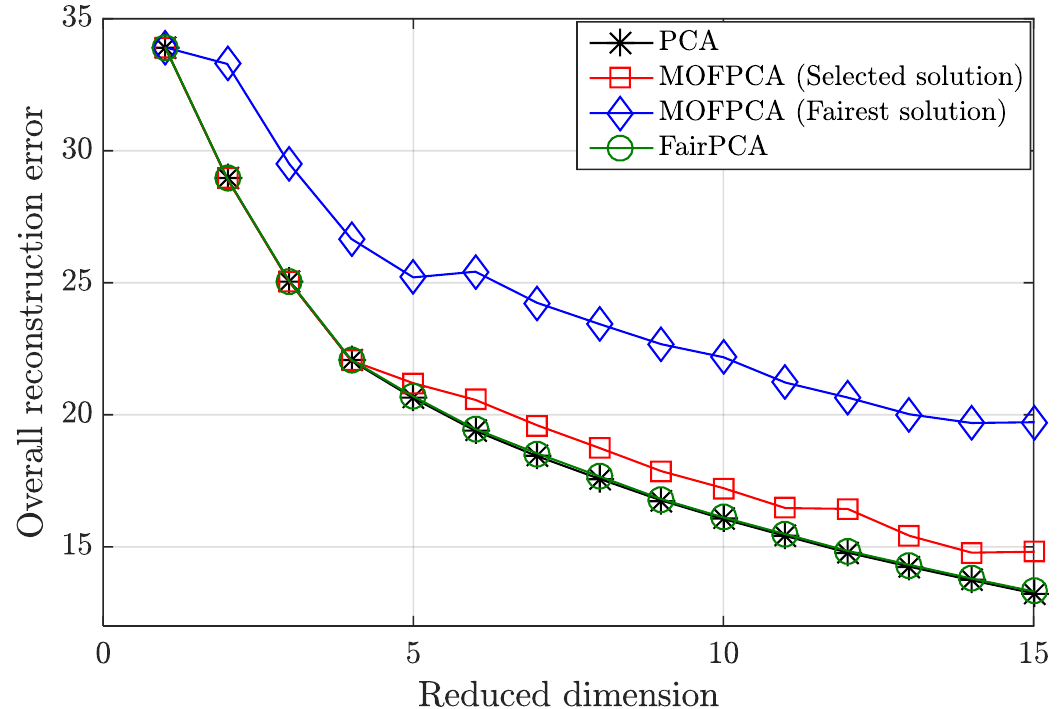}
\label{fig:result_recerror_images}}
\hfil
\subfloat[Fairness measure.]{\includegraphics[width=3.0in]{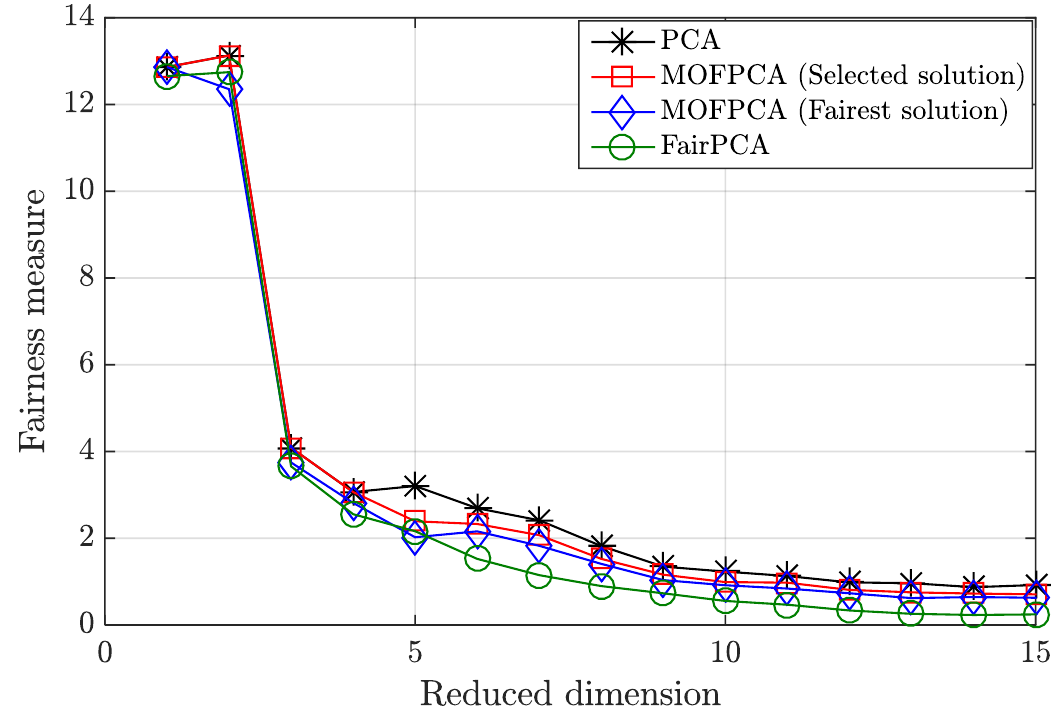}
\label{fig:result_fairmeas_images}}
\caption{Results for both cost functions and different numbers of reduced dimensions - LFW dataset.}
\label{fig:result_all_images}
\end{figure}

\begin{figure*}[!h]
\centering
\subfloat[PCA.]{\includegraphics[width=1.6in]{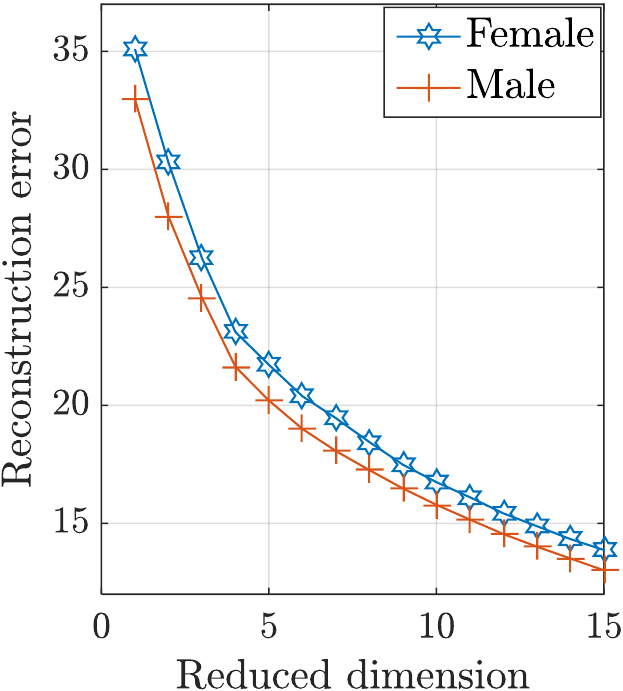}
\label{fig:reconserror_pca_images}}
\hfil
\subfloat[MOFPCA (Selected solution).]{\includegraphics[width=1.6in]{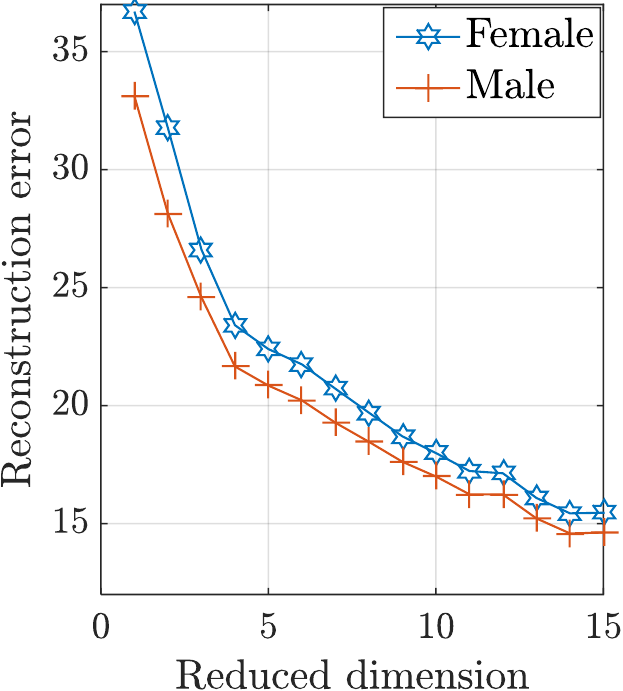}
\label{fig:reconserror_mofpca_images}}
\hfil
\subfloat[MOFPCA (Fairest solution).]{\includegraphics[width=1.6in]{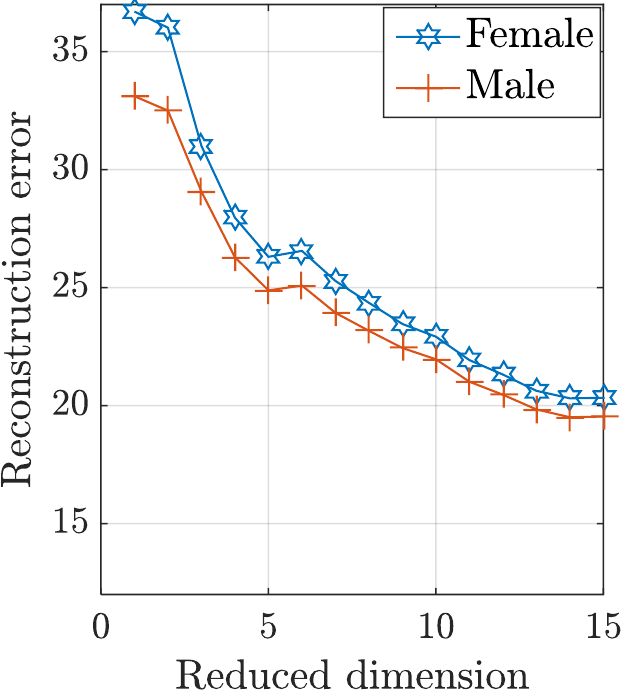}
\label{fig:reconserror_mofpca_f_images}}
\hfil
\subfloat[FairPCA.]{\includegraphics[width=1.6in]{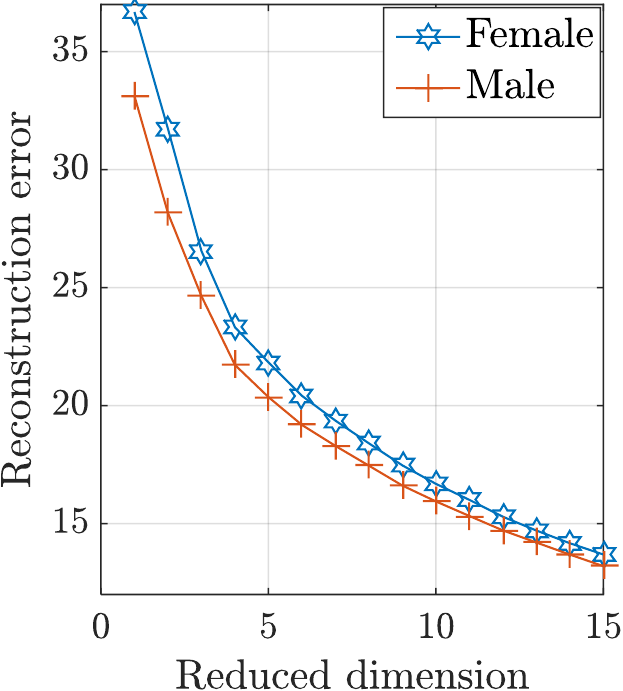}
\label{fig:reconserror_samadi_images}}
\caption{Reconstruction errors for females males - LFW dataset.}
\label{fig:reconserror_all_images}
\end{figure*}

Figure~\ref{fig:result_all_images} presents the obtained reconstruction error and fairness measure for different numbers of reduced dimensions. Although MOFPCA achieved better values of fairness measure in comparison with PCA, in this experiment, the FairPCA led to the better results for $r > 2$ (or $r > 4$ if one considers the fairest solution). Figure~\ref{fig:reconserror_all_images} illustrates the difference between the reconstruction errors for each class and each approach. One may note that the reconstruction errors provided by the FairPCA (Figure~\ref{fig:reconserror_samadi_images}) are closer in comparison with PCA and MOFPCA. A hypotheses for this result is that, since the LFW dataset comprises 1764 attributes (in contrast with the 22 in the Default Credit dataset) and the FairPCA search for projection vectors different from the classical PCA, this method has more flexibility to adjust them in order to enhance fairness.

\subsection{Experiments with the LFW dataset and balanced samples}
\label{subsec:exper3}

\begin{figure}[ht]
\centering
\subfloat[Reconstruction error.]{\includegraphics[width=3.0in]{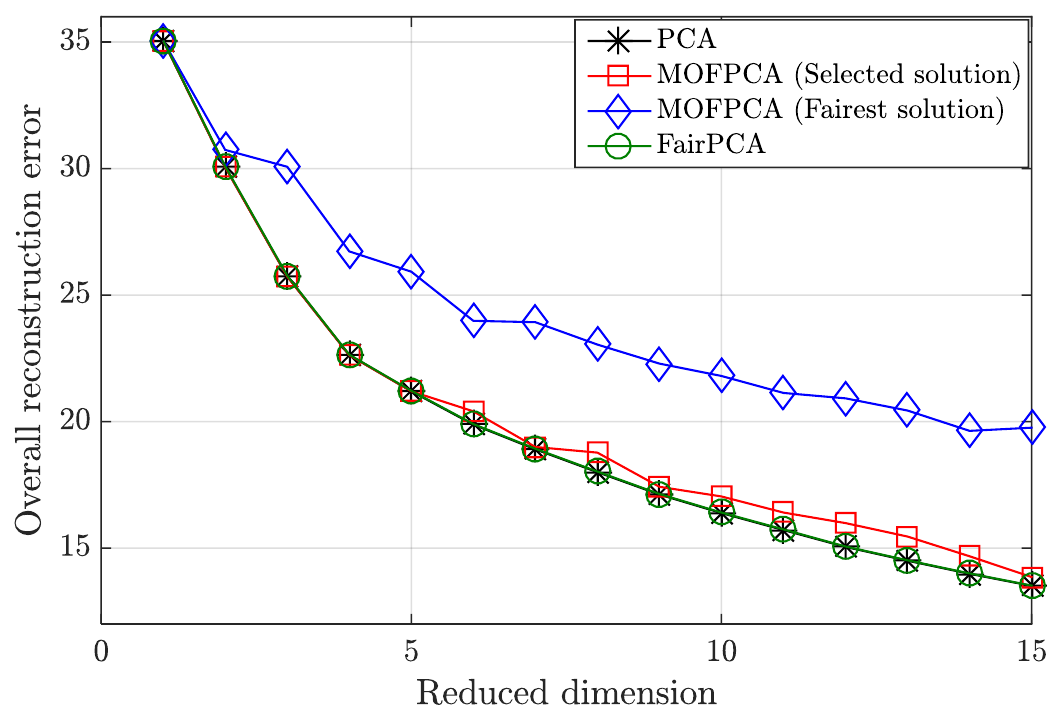}
\label{fig:result_recerror_images_eq}}
\hfil
\subfloat[Fairness measure.]{\includegraphics[width=3.0in]{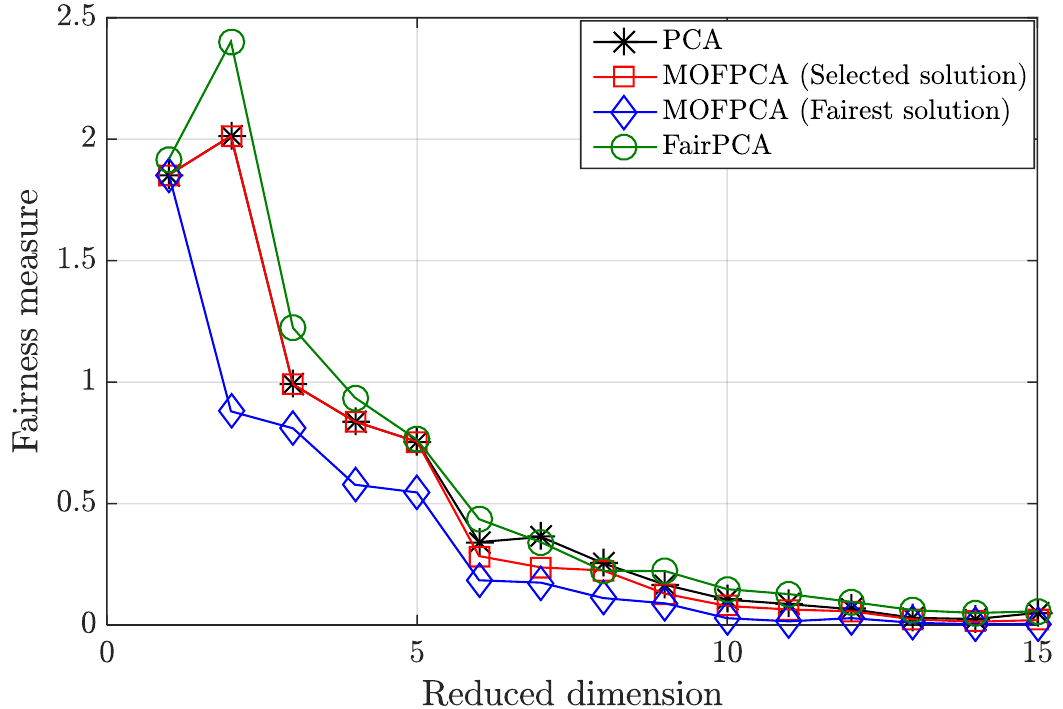}
\label{fig:result_fairmeas_images_eq}}
\caption{Results for both cost functions and different numbers of reduced dimensions - LFW dataset with balanced samples.}
\label{fig:result_all_images_eq}
\end{figure}

\begin{figure*}[!h]
\centering
\subfloat[PCA.]{\includegraphics[width=1.6in]{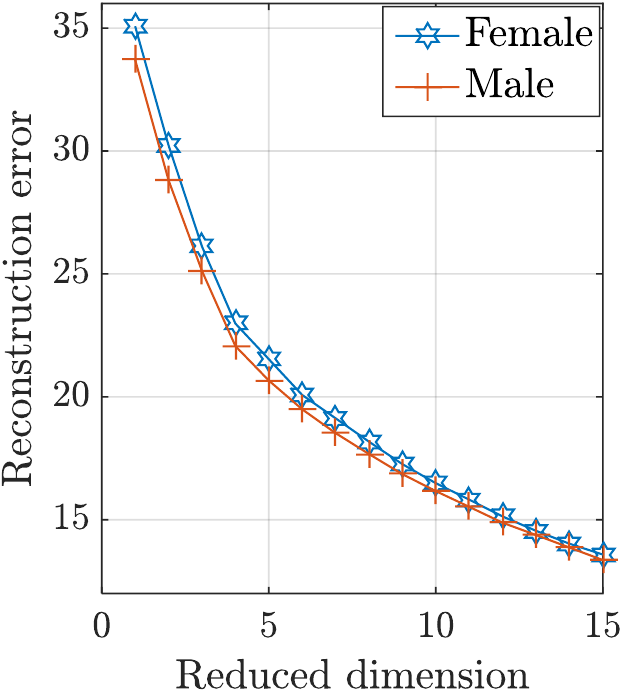}
\label{fig:reconserror_pca_images_eq}}
\hfil
\subfloat[MOFPCA (Selected solution).]{\includegraphics[width=1.6in]{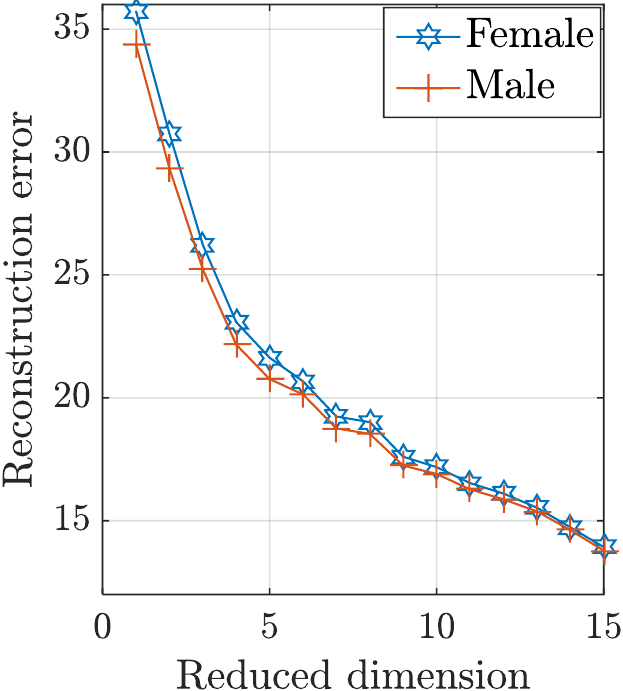}
\label{fig:reconserror_mofpca_images_eq}}
\hfil
\subfloat[MOFPCA (Fairest solution).]{\includegraphics[width=1.6in]{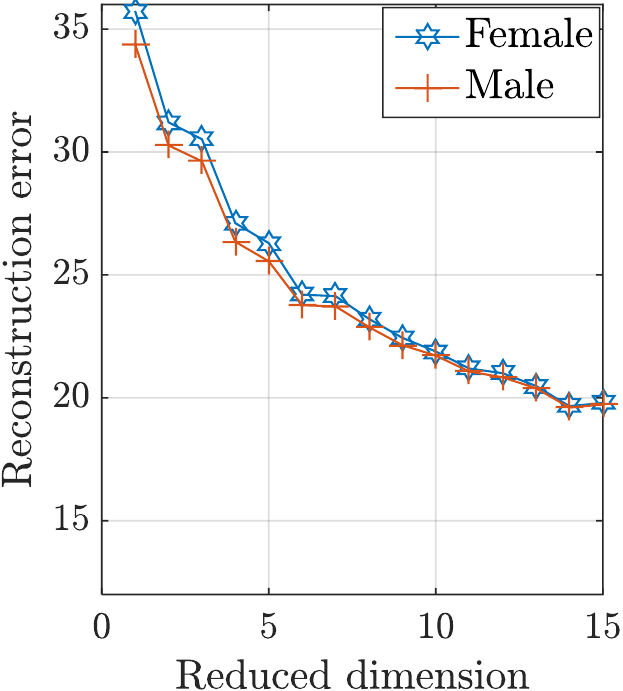}
\label{fig:reconserror_mofpca_f_images_eq}}
\hfil
\subfloat[FairPCA.]{\includegraphics[width=1.6in]{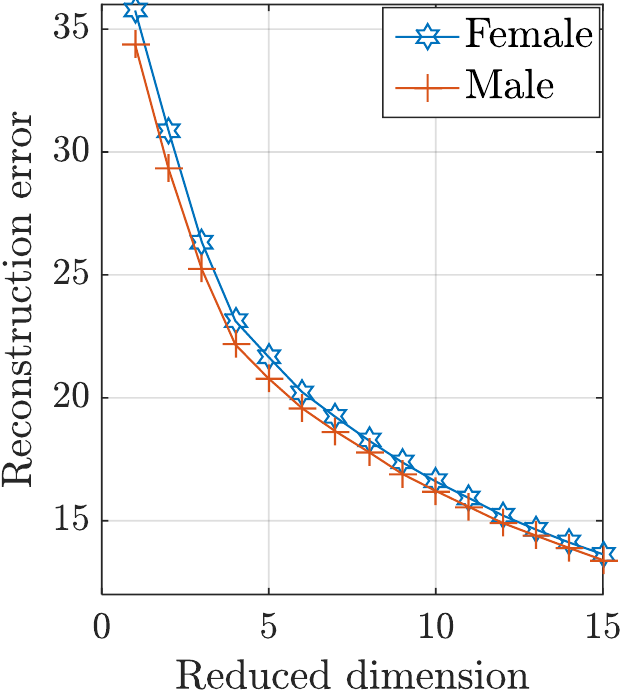}
\label{fig:reconserror_samadi_images_eq}}
\caption{Reconstruction errors for females males - LFW dataset with balanced samples.}
\label{fig:reconserror_all_images_eq}
\end{figure*}

Aiming at further investigating the application of our proposal in the LFW dataset, we consider a scenario with balanced samples with respect to the sensitive attribute. Therefore, we selected $n = 5,924$ samples divided into $n_1=2,962$ females and $n_2=10,270$ males. The results are presented in Figures~\ref{fig:result_all_images_eq} and~\ref{fig:reconserror_all_images_eq}. As can be seen in Figure~\ref{fig:result_recerror_images_eq}, the MOFPCA selected solution led to the overall reconstruction error close to both PCA and FairPCA solutions. However, in contrast with the previous experiment, it led to slightly better values of fairness measure (see Figure~\ref{fig:result_fairmeas_images_eq}), specially for $r > 6$. Therefore, even in a scenario with balanced samples, the proposed MOFPCA approach could improve fairness with some loss in the reconstruction error. The FairPCA could not reduce the disparity between the reconstruction errors and achieved a performance close to the classical PCA (see Figures~\ref{fig:reconserror_pca_images_eq} and~\ref{fig:reconserror_samadi_images_eq}). 

\section{Conclusions}
\label{sec:concl}

Ethical concerns in AI have become an important subject in the last years. Automatic decision systems should take into account fairness in order to avoid disparate treatment of different sensitive groups. For instance, in dimensionality reduction problems, one should adopt a procedure that provides an equal (or, at least, as similar as possible) representation of different groups. In this context, this paper proposed a multi-objective framework for the Fair Principal Component Analysis. Our approach consists in a different ordering of the components given by the classical PCA, which gives rise to an approach with low computational costs and easy to deploy in already running systems. Furthermore, the set of non-dominated solutions provides a suitable portfolio of choices to the stakeholders and decision makers, presenting, clearly, the trade-off between the objectives. This characteristic is paramount in scenarios of social and economic impacts, since the system user can measure the gain in terms of fairness with a (possible) small loss in the reconstruction error.

We verified the applicability of our proposal in experiments based on two datasets frequently used in the literature. In the Default Credit dataset, we could attest that there is a non-dominated solution that lead to very good values in both reconstruction error and fairness measure. The disparity between the representation of the two groups was considerably reduced with an small loss in the reconstruction error. Moreover, the proposed MOFPCA approach performed much better in comparison with both classical PCA and FairPCA. However, the experiments with the LFW dataset indicated that our proposal may lead to less expressive results when the number of attributes are very high. Even when the samples are balanced with respect to the sensitive attributes, there were a slightly gain with the MOFPCA approach in comparison with PCA or FairPCA. In high-dimensional datasets, we see that future works could be developed in order to generalize our proposal to the search of any projection matrix (not necessarily based on the classical PCA) while using the multi-objective framework. Therefore, one could either achieve better solutions, in terms of minimizing both objectives, and allow the trade-off analysis among the non-dominated set.

The scope of this paper lies in the fairness analysis in the preprocessing step. Note that we do not address, necessarily, a classification problem. Therefore, as another future perspective, we would like to verify the MOFPCA impact on the classification task. More specifically, we intend to investigate if looking at fairness in the preprocessing step lead to fairness in the classification task. Moreover, new researches in classification problems may consider a multi-objective formulation in which the cost functions are associated with model accuracy and fairness. Although our proposal tackled fairness in preprocessing step, it could be generalized to analyze the trade-offs in any step of a machine learning problem.

\bibliographystyle{IEEEtran}
\bibliography{IEEEabrv,_ref_wcci2022}

\end{document}